\documentclass[lettersize,journal]{IEEEtran}
\usepackage{amsmath,amsfonts}
\usepackage{algorithmic}
\usepackage{algorithm}
\usepackage{array}
\usepackage{textcomp}
\usepackage{stfloats}
\usepackage{url}
\usepackage{verbatim}
\usepackage{graphicx}
\usepackage{cite}
\usepackage{booktabs}
\usepackage{adjustbox}
\usepackage{subcaption}
\usepackage{caption}
\usepackage{xcolor}
\usepackage{hyperref}
\usepackage{siunitx}
\usepackage{placeins} 

\begin{document}

\title{
A Neuromorphic Incipient Slip Detection System using Papillae Morphology
}

\author{Yanhui Lu, 
        Zeyu Deng, 
        Stephen J. Redmond,~\IEEEmembership{Senior Member,~IEEE}, 
        Efi Psomopoulou,~\IEEEmembership{Member,~IEEE}, 
        Benjamin Ward-Cherrier,~\IEEEmembership{Member,~IEEE}%
\thanks{This work was partially supported by the China Scholarship Council, the Horizon Europe research and innovation program (MANiBOT, Grant 101120823), the Royal Society International Collaboration Awards (South Korea), ARIA on Robot Dexterity and the Royal Academy of Engineering Fellowship (Grant RF02021071).}%
\thanks{Y. Lu, E. Psomopoulou, and B. Ward-Cherrier are with the School of Engineering Mathematics and Technology, University of Bristol, U.K.; Z. Deng is with the James Watt School of Engineering, University of Glasgow, U.K.; S. J. Redmond is with the School of Electrical and Electronic Engineering, University College Dublin, Ireland.}%
\thanks{Corresponding author: Yanhui Lu (email: mb23300@bristol.ac.uk).}%
}
% The authors are with the School of Engineering Mathematics and Technology, 
% University of Bristol, Bristol, U.K. (e-mail: mb23300@bristol.ac.uk;
% b.ward-cherrier@bristol.ac.uk; efi.psomopoulou@bristol.ac.uk); 
% the James Watt School of Engineering, University of Glasgow, Glasgow, U.K. 
% (e-mail: 3139275D@student.gla.ac.uk); and the School of Electrical and 
% Electronic Engineering, University College Dublin, Dublin, Ireland 
% (e-mail: stephen.redmond@ucd.ie).}}%


% Remember, if you use this you must zcall \IEEEpubidadjcol in the second

\maketitle

\bstctlcite{myIEEEctl}

\begin{abstract}

Detecting incipient slip enables early intervention to prevent object slippage and enhance robotic manipulation safety. However, deploying such systems on edge platforms remains challenging, particularly due to energy constraints. This work presents a neuromorphic tactile sensing system based on the NeuroTac sensor with an extruding papillae-based skin and a spiking convolutional neural network (SCNN) for slip-state classification. The SCNN model achieves \text{94.33\%} classification accuracy across three classes (no slip, incipient slip, and gross slip) in slip conditions induced by sensor motion. Under the dynamic gravity-induced slip validation conditions, after temporal smoothing of the SCNN’s final-layer spike counts, the system detects incipient slip at least \text{360 ms} prior to gross slip across all trials, consistently identifying incipient slip before gross slip occurs. These results demonstrate that this neuromorphic system has stable and responsive incipient slip detection capability.

\end{abstract}

% \begin{IEEEkeywords}
% tactile sensing, texture classification, force modulation, dynamic touch, robotic perception, neuromorphic
% \end{IEEEkeywords}

\vspace{-1.5em}

\section{Introduction}

When humans grasp objects, mechanoreceptors beneath the fingertip skin encode relative motion between the skin and object, converting it into electrical signals for rapid, precise neural responses. This mechanism is thought to enable humans to detect slip at the fingertip and may support tactile-guided grasp control. \cite{johansson1984roles}. Similarly, drawing on tactile cues observed in humans, robotic systems may better adapt to dynamic grasping in complex environments \cite{james2018slip,hou2024piezoresistive}.

Slip detection is typically divided into gross slip, defined as complete sliding between contact surfaces \cite{adams2013finger}, and incipient slip, the early stage where part of the contact area moves while the rest remains stationary \cite{afzal2024role}. Relying only on gross slip detection risks delayed adjustment and object drop. Therefore, incipient slip detection facilitates earlier corrective measures that enhance the safety of robotic manipulation. \cite{chen2018tactile}.

\begin{figure}[!t]
    \centering
    \includegraphics[width=0.8\linewidth]{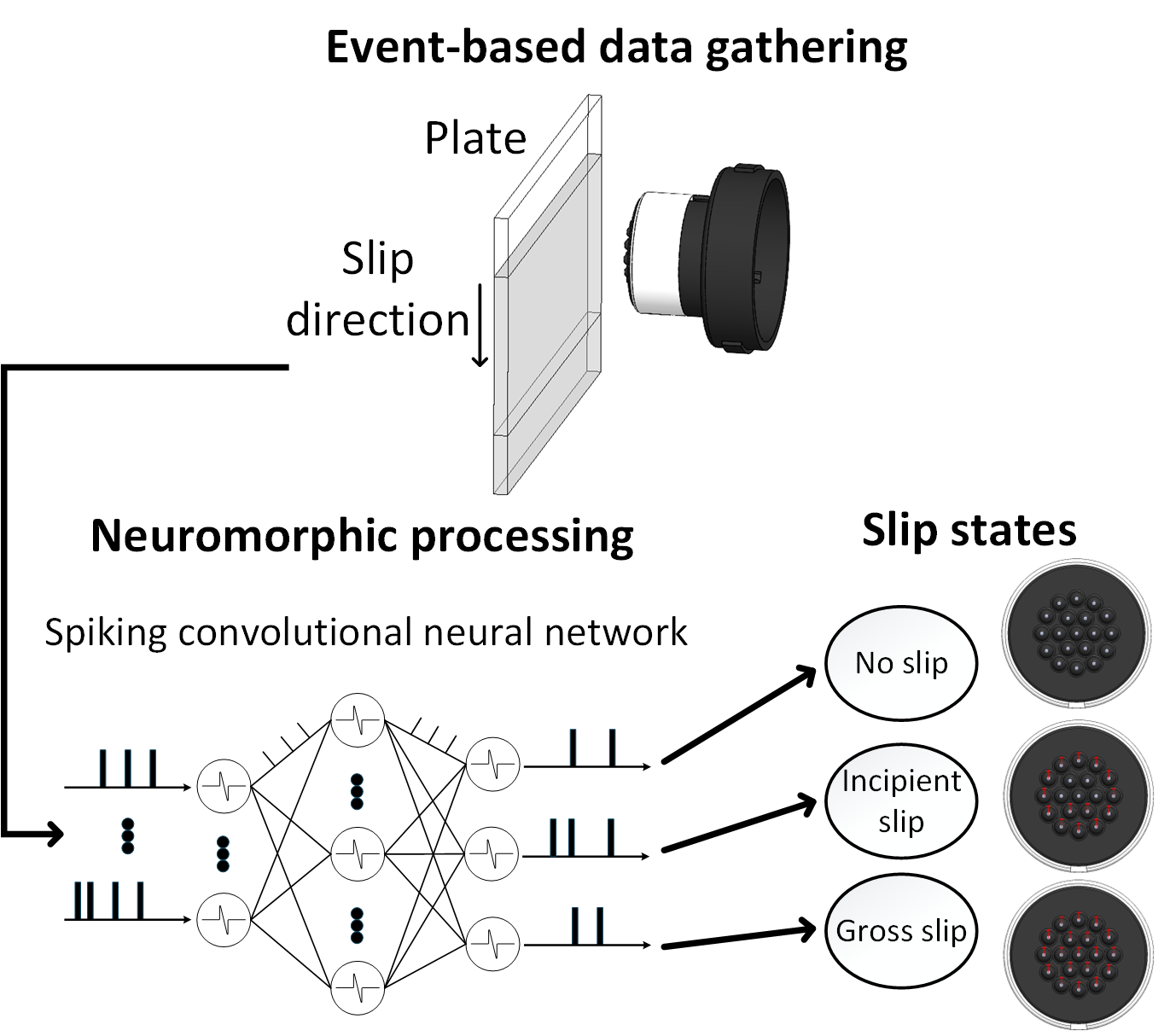}
    \caption{The proposed slip state detection system. After acquisition by the sensor equipped with a novel skin, the event-based data is preprocessed and fed into a spiking convolutional neural network (SCNN), followed by temporal smoothing of the final layer's spike counts to detect the slip state.}
    \label{fig:whole}
\end{figure}

In recent years, vision-based tactile sensors (VBTS) have enabled effective tactile perception across diverse contact conditions, owing to their high sensitivity and spatial resolution \cite{li2024vision}. However, the fine temporal resolution needed for slip detection, together with the high energy cost of frame-based data acquisition and computation make conventional VBTS less optimal. Given these constraints, neuromorphic vision-based tactile sensors (NVBTS) \cite{ward2020neurotac,funk2024evetac} offer promising solutions to these limitations. NVBTS typically employ event cameras, which operate asynchronously and generate events only upon significant local brightness changes, each tagged with a polarity indicating brightness increase (positive) or decrease (negative) \cite{gallego2020event}. They offer greater dynamic precision while avoiding the redundant energy and memory demands of frame-wise acquisition \cite{chakravarthi2024recent}. Moreover, by integrating NVBTS with neuromorphic hardware \cite{orchard2021efficient,liu2019live}, which is optimized for asynchronous, event-driven execution of spiking neural networks (SNNs), the overall architecture can further reduce power consumption and latency in tactile data processing \cite{tenzin2024application}.

Thus, the neuromorphic slip detection pipeline holds promise for edge robotic applications, where fast and low-power tactile control is critical. Its event-driven nature also aligns with neural-inspired architectures \cite{zaghloul2006silicon}, enabling potential integration with human neural systems.

In this paper, we present an end-to-end neuromorphic system for incipient slip detection (Fig.~\ref{fig:whole}). Our contributions are as follows:

\begin{itemize}
  \item A novel application of a papillae-based soft skin for NVBTS, featuring a concentric pillar array that facilitates the induction of incipient slip;
  \item A fully neuromorphic incipient slip detection pipeline based on an SCNN applied to a neuromorphic vision-based tactile sensor;
  \item An evaluation of the method's detection latency under both kinematically controlled settings and dynamic real-environment validation tasks.
\end{itemize}

\section{Related work}

Existing studies have investigated the biomechanical characteristics of human skin during incipient slip. Findings indicate that the hemispherical structure of the human fingertip induces spatial variations in pressure \cite{birznieks2001encoding}, and, in combination with structures such as fingerprint ridges, contributes to spatially heterogeneous shear deformation across the contact area \cite{corniani2023sub}. This facilitates the progressive propagation of slip, typically from the periphery toward the center of the contact area. These insights not only enhance our understanding of human tactile perception but also provide inspiration for designing artificial tactile sensors capable of mimicking skin incipient slip behavior.

In incipient slip detection research, smooth planar or hemispherical sensor surfaces are commonly adopted \cite{su2015force,yuan2015measurement,sui2022incipient}. However, such designs often hinder shear deformation of the skin and thus fail to effectively induce authentic incipient slip \cite{bulens2023incipient}. Furthermore, the observed slip phenomena in these studies are often difficult to attribute directly to incipient slip, due to factors like inconsistent definitions and indirect measurement methods. Inspired by the skin mechanism during incipient slip, some researchers have explored various structural designs to actively induce this phenomenon. For example, certain studies used a \num{3}$\times$\num{3} optical papillary array, where the central papilla is elevated to delay its slip response \cite{khamis2018papillarray,ulloa2022incipient,wang2023robust}. Others enhanced the independence of local slip regions by modifying the skin layer of VBTS, such as adding concentric fingerprint-like ridges \cite{james2020biomimetic} or rigidly connected outer ridges \cite{bulens2023incipient}. In this work, we propose an improved papillary skin structure for NVBTS that induces clearer and more localized surface slip phenomena, thereby facilitating neuromorphic incipient slip detection.

In terms of incipient slip detection methods, most studies place greater emphasis on physics-based models rather than relying on data-driven learning approaches. For example, one study filtered the pressure signal from a BioTac sensor to estimate vibration energy, enabling the detection of incipient slip \cite{su2015force}. Additionally, a DAVIS event camera was employed to analyze variations in the contact area through positive and negative events to identify the start of incipient slip \cite{rigi2018novel}. Another work leveraged a GelSight tactile sensor to track the entropy of a randomly distributed marker displacement field on the elastic surface for incipient slip detection \cite{yuan2015measurement}. Sui et al.\ employed a VBTS, constructing a linear relationship between the distributed force and deformation gradient to quantify the extent of incipient slip \cite{sui2022incipient}. Moreover, incipient slip detection was also achieved in the PapillArray sensor by analyzing parameters such as papillae spatial deformation and contact force profiles, in combination with a velocity threshold criterion \cite{ulloa2022incipient}. These methods are often constrained by environmental conditions, making the results susceptible to external interference. Moreover, as the complexity of the contact interface increases, more sophisticated quantification methods are typically required, which adds significant challenges to the detection process.

In learning-based approaches, existing studies often rely on independent external cameras to capture markers on the outer surface of a tactile sensor, enabling the identification of the actual timing of slip events to generate ground truth labels. These labels are subsequently used to train models for incipient slip detection. One such report integrated multiple frames captured by an internal camera and fed the merged sequence into a convolutional neural network for detection \cite{james2020biomimetic}. Another approach extracted normalized velocity vectors by tracking the displacement of internal pins across consecutive frames, using these vectors as input to machine learning models for incipient slip recognition \cite{bulens2023incipient}. A further study employed the PapillArray sensor, applying tactile data augmentation and using the XY-plane velocity of the sensor’s papillae as input to a GRU-based recurrent neural network, achieving high-accuracy incipient slip detection through a multi-model prediction strategy \cite{wang2023robust}.

Although data-driven approaches have addressed several limitations of physics-based methods, conventional learning models running on GPU or CPU platforms remain resource-intensive, particularly in real-world robotic environments where budgets for power, weight, and other resources are tightly constrained. To address this, we introduce a neuromorphic approach using a Spiking Convolutional Neural Network (SCNN) to enhance generalization in incipient slip detection while reducing resource demands.

\section{Methods}

\subsection{Sensor structure design}

\begin{figure}[!t]
    \centering
    \includegraphics[width=\linewidth]{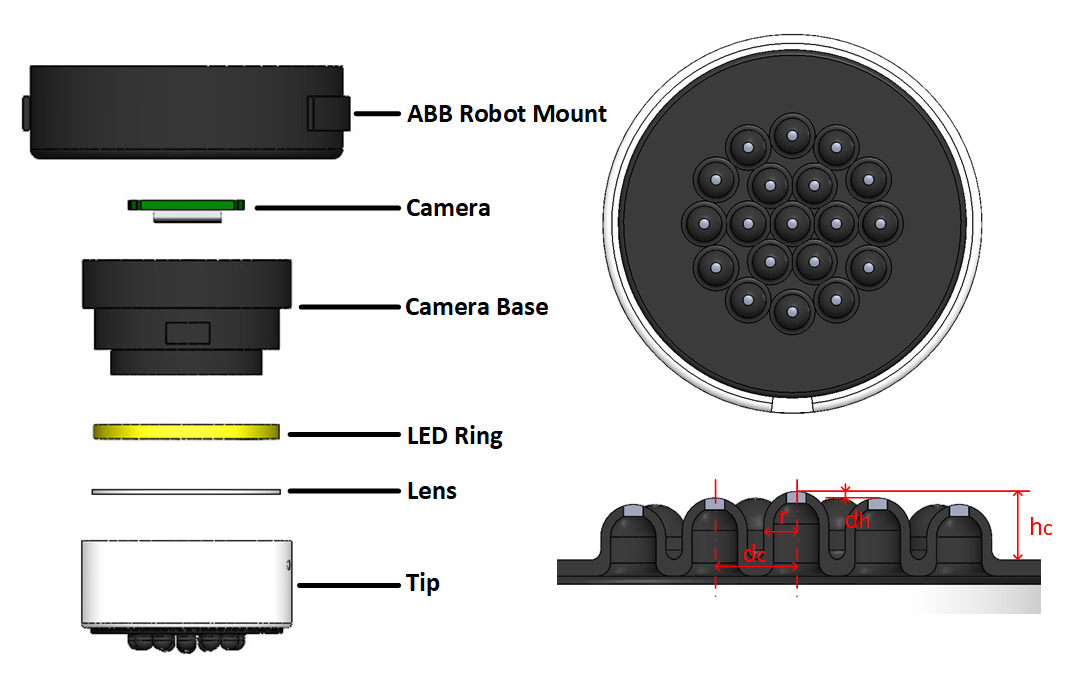}
    \caption{Overview of the new sensor structure. The left panel illustrates an exploded view of the sensor components, while the right panel shows the top view of the skin’s concentric papilla layout, along with a lateral cross-sectional view.}
    \label{fig:structure}
\end{figure}

Building on prior findings that the smooth hemispherical skin of the original NeuroTac sensor limits partial shear deformation for incipient slip due to strong mechanical coupling between skin regions \cite{james2020biomimetic}, we propose a newly designed NVBTS skin, as illustrated in Fig.~\ref{fig:structure}. The skin comprises three concentric layers of externally protruding papillae, with the papilla height gradually decreasing from the center to the outermost layer, aiming to induce earlier slip in the outer papillae. To prevent interference with independent papilla motion, the internal pins of the original skin were removed, allowing the internal event camera to directly capture events induced by papilla deformation for slip detection. Additionally, the central area on top of the papillae was printed using white material to facilitate external visual tracking of the papilla slip behavior for reference.

During sensor fabrication, silicone was injected into the tip cavity and cured in place, slightly bulging the soft outer skin due to internal pressure and causing minor deviations in papilla height and position. In this study, post-fabrication measurements showed the central papilla height above the original skin plane $\mathrm{h_c} = 10.1\,\text{mm}$. The height difference between adjacent layers of papillae $\mathrm{d_h} = 1.1\,\text{mm}$, and the spacing distance between adjacent layers of papillae $\mathrm{d_c} = 5.6\,\text{mm}$. The radius of each papilla $\mathrm{r} = 2.0\,\text{mm}$.

\subsection{Data collection and preprocessing}

\subsubsection{Kinematic-controlled experiments}

\begin{figure}[!t]
    \centering
    \includegraphics[width=0.8\linewidth]{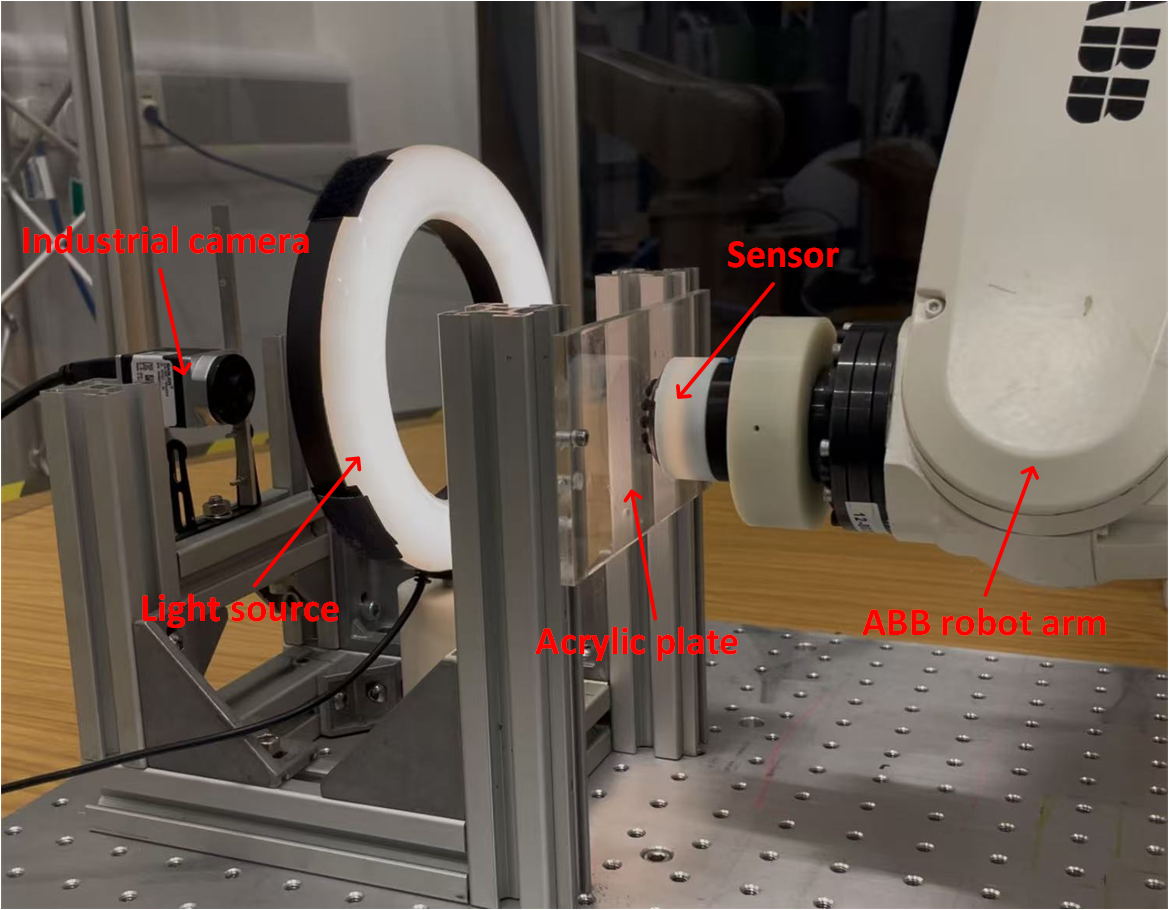}
    \caption{Setup for data collection in kinematically controlled environments. Sensor mounted on an ABB IRB120 robotic arm; transparent acrylic plate vertically mounted on a metal frame; external industrial camera fixed opposite to the contact surface for reference annotation.}
    \label{fig:setup}
\end{figure}

The data collected in this experiment, depicted in Fig.~\ref{fig:setup}, was used for model training. A 6-DOF robotic arm served as the actuator (ABB, IRB120), with a novel NeuroTac tactile sensor mounted as its end-effector. During collection, an acrylic plate was vertically mounted to a metal frame, and a 100 fps camera (Basler a2A1920-160ucBAS) captured the external displacements of the sensor papillae, forming our ground truth slip state.

Each sliding trial was executed with fixed indentation depth ($2.4\,\text{mm}$ to $3.4\,\text{mm}$ in $0.2\,\text{mm}$ increments) and constant sliding speed ($0.6\,\text{mm/s}$ to $1.6\,\text{mm/s}$ in $0.2\,\text{mm/s}$ steps). 
%to ensure full contact across all pillar layers and to cover a representative range of contact pressures, while preventing excessive deformation that could lead to mechanical wear. To improve detection performance under varying speeds, we used sliding speeds to capture a representative spectrum of event rates generated by the sensor.
For each combination of indentation depth and sliding speed, the sensor was moved tangentially along the plate surface over a distance of $15\,\text{mm}$ in one of eight directions, spaced at $45^{\circ}$ intervals from $0^{\circ}$ to $315^{\circ}$, to induce different slip states. Each direction was repeated three times, resulting in a total of 864 sliding trials.

\subsubsection{Gravity-induced dynamic experiments}

To evaluate the model’s ability to detect slip states under dynamic and realistic conditions, we designed two gravity-induced experiments. In these experiments, the robotic arm pressed the sensor against an raised acrylic plate, held still, and then retracted slowly. The plate gradually slipped once the friction force reduced below the plate's weight, sequentially exhibiting first incipient and then gross slip. %These setups introduced dynamic changes in both depth and speed, along with slight vibrations caused by looseness in the slider attachment.

Firstly, to ensure a smooth and sufficiently long sliding process for capturing the incipient slip phase, while keep the tangential force within the supportable range of the sensor’s indentation depth, we conducted a factorial experiment with three plate weights ($0.165\,\text{kg}$, $0.205\,\text{kg}$, and $0.245\,\text{kg}$) and three retraction speeds ($0.3\,\text{mm/s}$, $0.5\,\text{mm/s}$, and $0.7\,\text{mm/s}$), yielding nine configurations to evaluate the model under varying contact conditions. For each configuration, 20 sliding trials were collected to systematically analyze the impact of plate weight and robot retraction speed on slip detection performance.

Subsequently, to evaluate the robustness of the model under slip trajectory perturbations, we conducted a second experiment using a $0.205\,\text{kg}$ plate weight and $0.5\,\text{mm/s}$ retraction speed as the baseline. During retraction, the robot simultaneously applied a horizontal motion to the sensor, perpendicular to the slip direction, at varying speeds: 25\%, 50\%, 75\%, and 100\% of the baseline retraction speed. For each disturbance level, 10 leftward and 10 rightward trials were collected, resulting in 20 trials per speed.

\FloatBarrier

\subsubsection{Data preprocessing}

To reduce computational load and response latency, only positive polarity events were used from the sensor camera. The raw sensor output (\num{640}$\times$\num{480}) was cropped to a \num{400}$\times$\num{400} pixel region centered on the active area, which was then downsampled via non-overlapping \num{20}$\times$\num{20} pixel pooling windows. Within each pooling window, event trains from all pixels were temporally aggregated by merging and sorting their event timestamps, resulting in a final spatial input resolution of \num{20}$\times$\num{20} for the neural network.

\subsection{Data generation and training}
\label{sec:yourlabel}

\begin{figure}[!t]
    \centering
    \includegraphics[width=\linewidth]{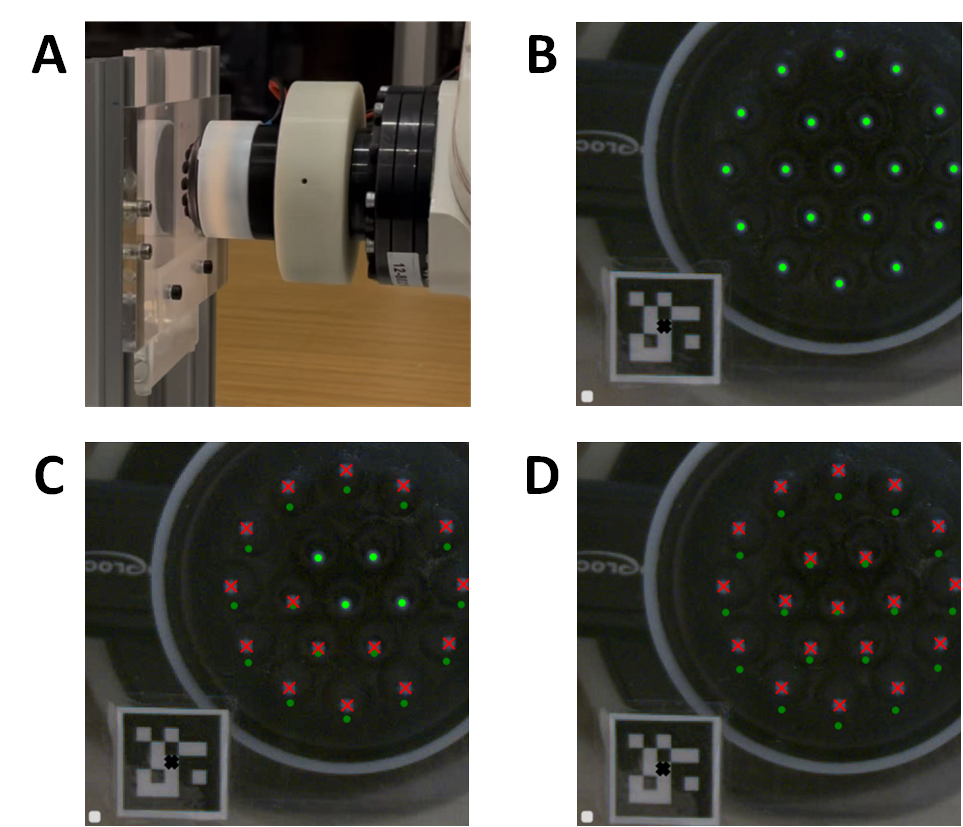}
    \caption{A) In gravity-induced experiments, the sensor is pressed against an acrylic plate with a sliding track and held stationary. B–D) Illustrations of the sensor–plate relative positions captured by an external camera during real-world experiments in the no slip, incipient slip, and gross slip phases. The green dots represent the initial relative positions of the white marks of the sensor papillae with respect to the center of the ArUco marker. Red markers indicate the papillae that have experienced slip according to a defined displacement threshold.}
    \label{fig:TEST_SETUP}
\end{figure}

The onset times of incipient and gross slip for each trial were initially annotated using external video recordings. White markers on the sensor papillae were tracked with OpenCV’s blob detection after thresholding. For each kinematically controlled trial, the displacement of each marker was calculated relative to its position in the first frame. A marker was considered slipping if its displacement exceeded a threshold of 2 pixels. For gravity-induced trials, marker slips were identified based on the relative displacement between the sensor's white markers and an ArUco marker on the acrylic plate, exceeding the same threshold \cite{bulens2023incipient}. The onset of incipient slip was defined as the first timestamp where any marker exceeded the displacement threshold (Fig.~\ref{fig:TEST_SETUP}C), while gross slip was defined as the timestamp where the central marker also crossed the displacement threshold (Fig.~\ref{fig:TEST_SETUP}D). 

For dataset extraction in training, the collected trials in our first experiment were split into training, validation, and test sets with a ratio of 70:15:15. For each trial, 50 consecutive $30\,\text{ms}$ samples were extracted from the no slip period before the onset of incipient slip, and another 50 samples were taken from the gross slip period following the end of incipient slip. 
%The window length was chosen to ensure sufficient event information for reliable detection while keeping the detection interval as short as possible.
For the incipient slip period, 50 non-overlapping $30\,\text{ms}$ samples were randomly selected if the duration was sufficient; otherwise, all available consecutive $30\,\text{ms}$ samples within the incipient slip interval were used. Each $30\,\text{ms}$ sample was split into 30 time steps, with positive-polarity events per step aggregated over a \num{20}$\times$\num{20} grid, yielding a \((30,\,1,\,20,\,20)\) tensor representing the spatiotemporal event segment.

For classification, we implemented a three-class SCNN using integrate-and-fire (IAF) neurons \cite{tal1997computing} from the Sinabs library \cite{sinabs}. The dynamics of IAF neurons can be described as follows:

\[
V_{\text{mem}}(t+1) = V_{\text{mem}}(t) + \sum_{i} z_i(t)
\]

\[
\text{if } V_{\text{mem}}(t) \geq V_{\text{th}}, \quad \text{then } V_{\text{mem}}(t) \leftarrow V_{\text{reset}}
\]

\noindent
where \( V_{\text{mem}}(t) \) denotes the membrane potential, \( V_{\text{th}} \) is the firing threshold, and \( V_{\text{reset}} \) is the reset potential after firing. \( \sum_{i} z_i(t) \) represents the sum of all input currents at time step \( t \).

\begin{figure}[!t]
    \centering
    \includegraphics[width=\linewidth]{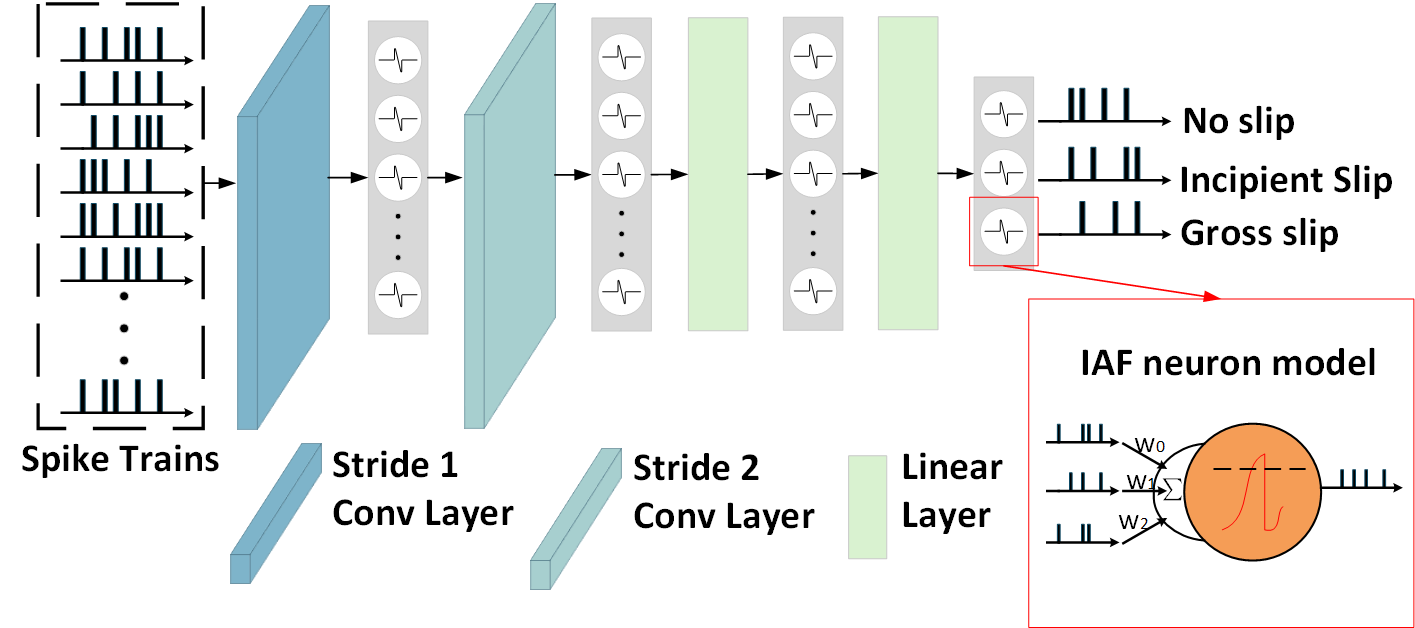}
    \caption{Schematic diagram of neural network architecture for slip state detection.}
    \label{fig:network}
\end{figure}

The network architecture consists of two convolutional layers followed by two fully-connected layers (Fig.~\ref{fig:network}). To reduce the spatial dimensions of feature maps after the second convolutional layer, we employed a strided convolution (stride of 2) instead of traditional pooling operations to enhance feature generalization while reducing computational cost. The final output layer contains three neurons, corresponding to the three slip state categories. 

The network is trained using surrogate gradient methods \cite{neftci2019surrogate}, and classification is determined based on the neuron in the output layer with the highest spike rate.

\subsection{Incipient and gross slip detection latency evaluation}

To evaluate the slip detection performance of the trained model, for each trial in both the kinematically controlled test sets and the gravity-induced experiments, consecutive $30\,\text{ms}$ data samples were extracted from the beginning of the trial to the point after gross slip, and processed into tensors matching the training data format. Each sample was sequentially fed into the trained network, with the model’s first estimation of the incipient slip class as the detected onset of incipient slip; thereafter, the first estimation of the gross slip class was recorded as the estimated gross slip onset.
%During the sliding process, the outer pillars typically exhibit incipient slip first, followed by gross slip involving all pillars. Therefore, the time when the model first predicts the incipient slip class is recorded as the estimated onset of incipient slip, after that the first prediction of the gross slip class is recorded as the estimated onset of gross slip. 
The detection latencies for incipient and gross slip are obtained by subtracting the ground truth times obtained in Section~\ref{sec:yourlabel} from the timestamps estimated by the classifier.

The model was selected based on classification accuracy using raw spike count outputs. A sliding window averaging (window length = 4) was then applied to the selected model to smooth its outputs and enhance detection stability, with the window length chosen based on the average incipient slip detection latency in the test set trials. This approach avoids the computational overhead of operations like softmax and better conforms to the computational principles of neuromorphic hardware. Furthermore, a detection was deemed valid only when the smoothed spike count for a class exceeded all others with a margin of at least 2, ensuring confident and stable detection.

%\vspace{-1em}

\section{Results}

\subsection{Data inspection}

\begin{figure}[!t]
    \centering
    \includegraphics[width=\linewidth]{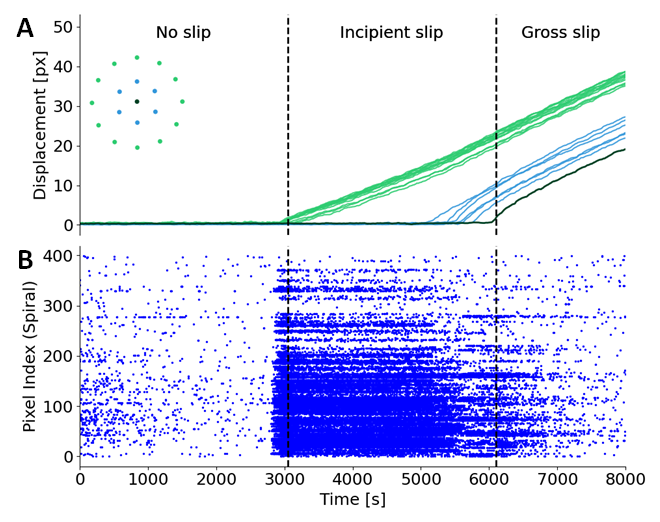}
    \caption{Displacement of white markers and corresponding event patterns during an example sliding trial. (A) The displacement of white markers on papilla tips of different sensor layers relative to their initial positions throughout the sliding trajectory. (B) Event-based activity recorded by the internal camera, where the vertical axis (Pixel Index) corresponds to an arrangement from the center outward, higher values indicate greater radial distance from the sensor center.}
    \label{fig:movement_train}
\end{figure}

\noindent
Just as incipient slip in human skin typically initiates at the periphery and propagates inward, the displacement trajectories of white markers on each papilla of the new skin exhibit a similar pattern, as demonstrated in a sample kinematically controlled trial (Fig.~\ref{fig:movement_train}A): the outer papillae slip first, followed by the central ones, indicating a progressive reduction in contact area prior to complete detachment. 

Meanwhile, Fig.~\ref{fig:movement_train}B shows the sensor’s event generation patterns over time. During the no slip phase, the event rate remains extremely low, primarily due to random sensor noise. Notably, events are mainly triggered by mismatches between the skin and camera movement. Before the onset of incipient slip, although the papillae have not yet started sliding, shear deformation begins to occur, leading to a sharp increase in the event rate. As more papillae gradually start to slip and their motion stabilizes, their shear deformation weakens and the event rate correspondingly decreases. After a short period into the gross slip phase, the sliding of all papillae becomes stable, and only a small number of localized events are generated by minor vibrations during the sliding process. This pattern indicates that events tend to be generated more frequently during the papilla "stick" phase than during the "slip" phase. Therefore, simple methods such as thresholding the event rate are insufficient to accurately detect the timing of incipient and gross slip, highlighting the need for more sophisticated approaches to model this dynamic process.

\subsection{Classification results and kinematic-controlled evaluation}

\begin{figure}[!t]
    \centering
    \includegraphics[width=0.9\linewidth]{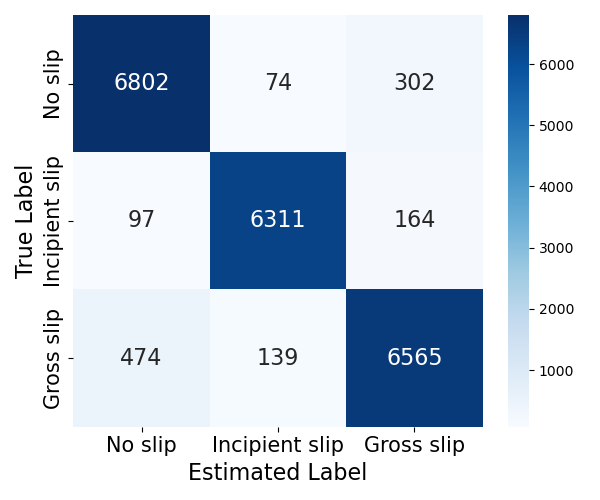}
    \caption{Confusion matrix of slip state classification results. Each classified sample corresponds to a processed tensor of size \((30,\,1,\,20,\,20)\) extracted from $30\,\text{ms}$ of neuromorphic tactile data.}
    \label{fig:confusion}
\end{figure}

\begin{figure}[!t]
    \centering
    \includegraphics[width=0.9\linewidth]{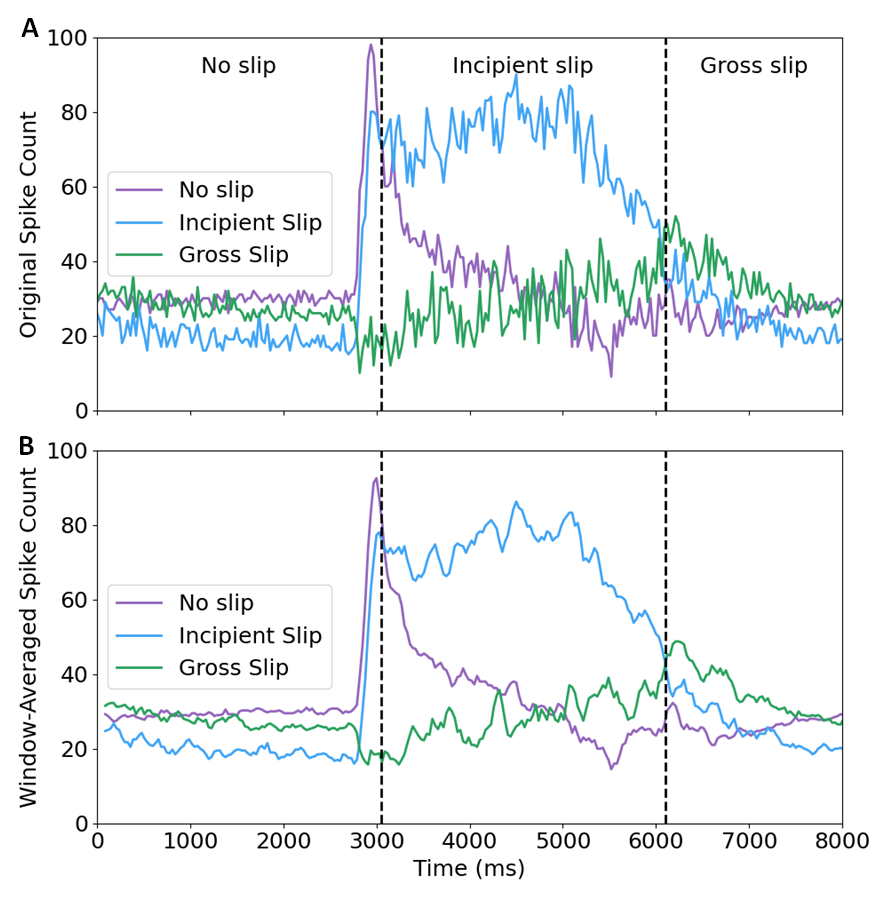}
    \caption{Comparison of model outputs before and after smoothing. (A) Raw spike count from the output neurons of the SCNN for each class over time. (B) Smoothed outputs using a windowed average, revealing clearer estimation transitions across slip stages.}
    \label{fig:output_compare}
\end{figure}

During the training phase, the best-performing model was selected based on validation set performance and subsequently evaluated on the test set. The final model, evaluated using its raw spike count outputs, achieved a classification accuracy of 94.33\% on the test set. The confusion matrix shown in Fig.~\ref{fig:confusion} indicates that the model effectively distinguishes between different slip states. Among them, the model achieved the highest precision (97.58\%) and recall (96.09\%) for the incipient slip class. In contrast, no slip and gross slip were more frequently confused, a pattern reflected in Fig.~\ref{fig:output_compare}, where the output spike counts appear highly similar during early no slip and late gross slip phases. This similarity may be explained by Fig.~\ref{fig:movement_train}A, where the sensor’s event rate drops sharply in late gross slip phase, resembling early no slip phase, which shows slightly higher activity than late no slip phase—possibly due to incomplete silicone recovery after compression.

Fig.~\ref{fig:output_compare} compares the raw model outputs with those smoothed using a temporal window. As shown, the unsmoothed outputs exhibit noticeable fluctuations and abrupt temporal changes across all three classes. In contrast, the smoothed outputs effectively suppress most of this variance, reducing its interference with detection and helping to avoid errors such as prematurely identifying incipient slip during the no slip phase.

\begin{figure}[!t]
    \centering
    \includegraphics[width=0.9\linewidth]{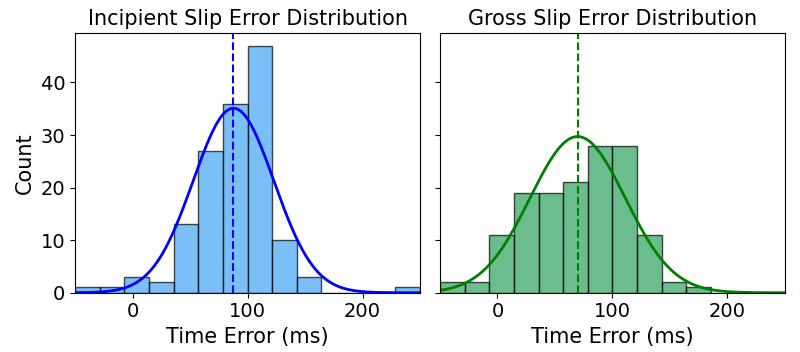}
    \caption{Incipient and gross slip detection timing latencies in kinematically controlled test environment.}
    \vspace{-10pt}
    \label{fig:structural}
\end{figure}

Building on the smoothed outputs, we further evaluated the model’s performance in terms of slip onset detection latency in kinematically controlled environment. On average, incipient slip was detected $88\,\text{ms}$ after its actual onset, while gross slip was detected on average $70\,\text{ms}$ after it occurs, with the distribution of their detection latencies presented in Fig.~\ref{fig:structural}.

\subsection{Evaluation in the gravity-induced experiments}

\begin{figure}[!t]
    \centering
    \begin{subfigure}[t]{\linewidth}
        \centering
        \includegraphics[width=\linewidth]{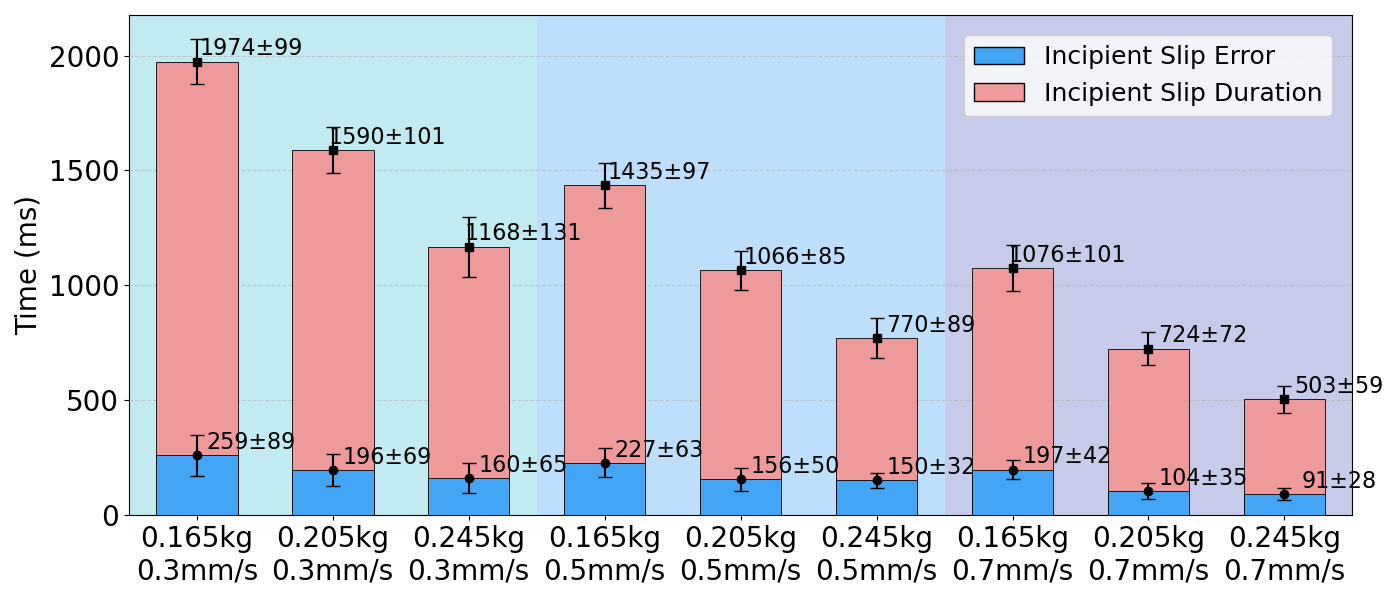}
        \caption{Incipient slip detection latencies under different condition combinations. Each bar shows the incipient slip duration (total height), with the lower portion representing the detection latency.}
        \label{fig:se}
    \end{subfigure}

    \vspace{0.8em} % 可选：用于在两个子图之间加一点垂直空隙

    \begin{subfigure}[t]{\linewidth}
        \centering
        \includegraphics[width=\linewidth]{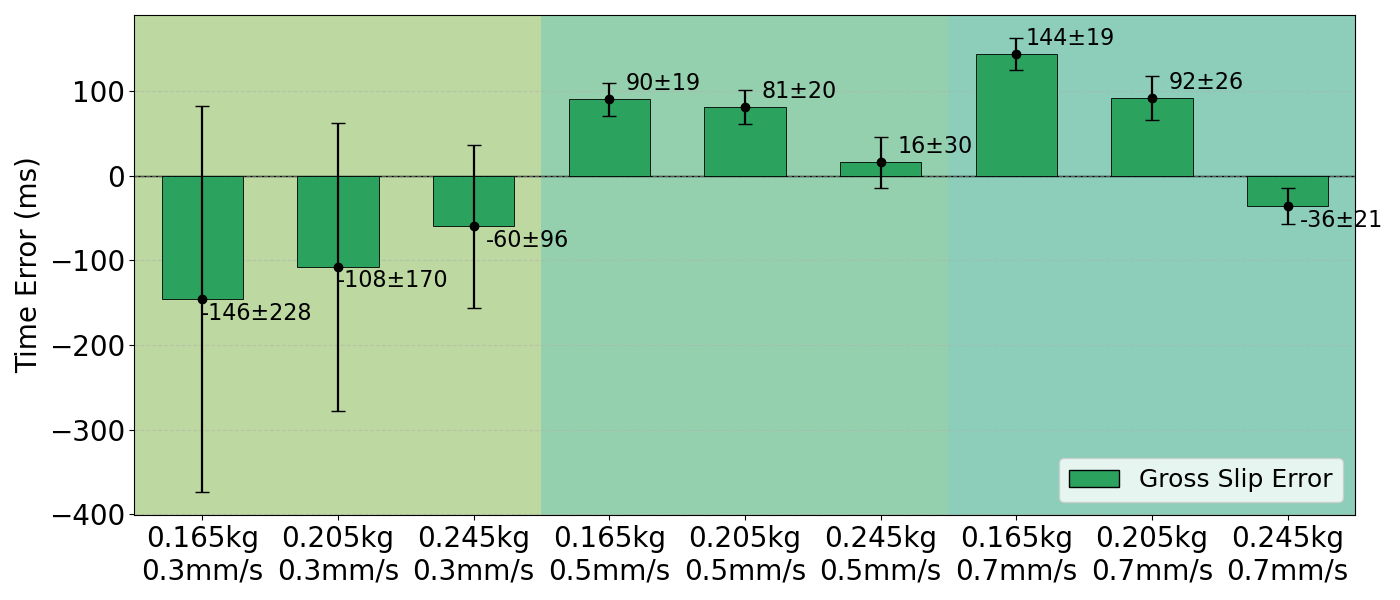}
        \caption{Gross slip detection latencies under different condition combinations.}
        \label{fig:ee}
    \end{subfigure}

    \caption{Model detection latencies across different combinations of plate weight and robot/sensor retraction speed.}
    \label{fig:te}
\end{figure}

We evaluated the model’s incipient slip detection latency across various combinations of applied plate weights and robot/sensor retraction speeds. Although arm retraction introduced a prolonged period of elevated event rates before slip onset (distinct from the no slip phase in the training data), Fig.~\ref{fig:te} shows that this condition had no significant impact on detection accuracy, suggesting that the model generalized well to such previously unseen conditions. Detection latencies were generally larger under lower retraction speeds or smaller plate weights, likely due to less continuous slip evolution at lower plate slip speeds, introducing more detection noise, or because such slow speeds were absent from the training data. Nevertheless, considering the observed duration of incipient slip under each condition, the model consistently detected incipient slip prior to the occurance of gross slip, indicating reliable early estimation across conditions.

In contrast to the trend observed for incipient slip, gross slip detection latencies did not vary systematically with weight or retraction speed. Under certain conditions (such as a low retraction speed $0.3\,\text{mm/s}$ or a large plate weight $0.245\,\text{kg}$ the model occasionally detected gross slip earlier than its actual occurrence, and the large standard deviation in detection latency observed at a retraction speed of $0.3\,\text{mm/s}$ also suggests high variability in its detection performance under this condition, indicating a certain degree of inconsistency in the model's performance for gross slip detection.

\begin{figure}[!t]
    \centering
    \includegraphics[width=\linewidth]{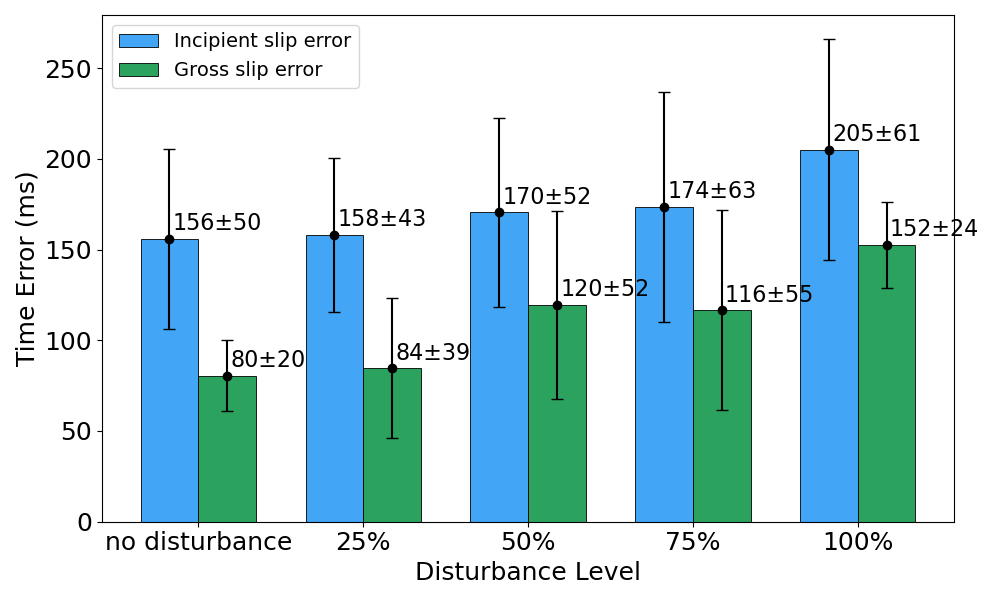}
    \caption{Incipient and gross slip detection latencies under different horizontal disturbance levels.}
    \label{fig:de}
\end{figure}

During the retraction phase of the robotic arm, when different levels of horizontal disturbances were introduced, for incipient slip, the detection latency slightly increased with larger disturbance speeds. Nonetheless, the model consistently detected incipient slip $691 \pm 118\,\text{ms}$ before the actual beginning of gross slip across all disturbance conditions. For gross slip, the detection latency generally increased with disturbance speed, although some minor fluctuations were observed. A summary of these trends across disturbance levels is shown in Fig.~\ref{fig:de}.

%When we further increased the disturbance magnitude to 125\% of the retraction speed, the model began to misclassify the onset of retraction as the onset of incipient slip. This larger disturbance caused more significant instantaneous shear deformation in the horizontal direction, which, when superimposed with the deformation along the slip direction, resulted in a complex composite deformation. Such deformation may interfere with the model’s ability to accurately detect the incipient slip state.

\section{Discussion}

This paper proposed a novel NeuroTac sensor skin with a concentric papilla array of varying heights to effectively induce incipient slip. Experimental results show that slip propagates from the outer papilla layer toward the center, resembling the partial slip response of the human fingertip.

Based on this, we integrated an event-driven learning method to accurately classify three slip states. In comparison to a previous study under similar training data collection settings that achieved 97.46\% accuracy under a single sliding speed \cite{bulens2023incipient}, our approach attained 94.33\% across multiple sliding speeds. With temporal smoothing of the model output, the system reliably detected the initiation of incipient slip across kinematically controlled experiments and multiple dynamic gravity-induced slip settings, with a minimum lead time of $360\,\text{ms}$ before the occurrence of gross slip in the gravity-induced settings across all trials, while retaining the ability to detect gross slip itself.

Compared to previous VBTS skins, with concentric ridges on hemispherical surfaces \cite{james2020biomimetic,bulens2023incipient}, the proposed papillae-based design further reduces mechanical coupling across regions of the skin and provides clearer slip timing between concentric rings of papillae, which facilitates slip margin estimation for better grasp strategy optimization. Using a soft material for the entire skin, however, can cause inconsistencies in geometric features during fabrication; adopting a rigid base with deformable papillae may mitigate this. Moreover, in contrast to earlier binary classification approaches that only distinguished incipient slip from non-incipient slip \cite{james2020biomimetic,wang2023robust}, our three-class model enables more fine-grained control, but still struggles to separate partial gross slip from static contact due to low event density during papilla steady slip, as both states may produce little papilla movement in the absence of texture or stick–slip events. Future work may explore structural designs that amplify micro-vibrations to highlight phase differences. In addition, evaluation should be extended to conditions not addressed here, such as curved or compliant surfaces, varying textures, rotational slips, and non-parallel contacts or motions.

%According to Ziegler et al.~\cite{ziegler2024detection}, inference on neuromorphic hardware (e.g., DynapCNN) offers power consumption several orders of magnitude lower than conventional GPUs. Moreover, direct integration of event cameras with neuromorphic processors eliminates bandwidth and latency bottlenecks in traditional architectures, enabling faster inference than GPU-based systems.
From a hardware deployment perspective, inference on neuromorphic hardware (e.g., DynapCNN) offers power consumption several orders of magnitude lower than conventional GPUs \cite{ziegler2024detection}, while direct integration of event cameras with neuromorphic processors eliminates the communication latency of neuromorphic computation, enabling faster inference than GPU-based systems \cite{ziegler2024detection}. These properties make our method highly promising for deployment in embedded tactile systems. Future work will focus on deploying the method on neuromorphic hardware to enable energy-efficient, rapid tactile sensing and control.

In conclusion, our approach provides an effective framework that combines morphological innovation and neuromorphic computation to achieve accurate and robust incipient slip detection, offering a promising direction toward real-time, low-power edge robotic applications.

\IEEEtriggeratref{28} % 比如 19 是触发点
\IEEEtriggercmd{\enlargethispage{-\baselineskip}}
\bibliographystyle{IEEEtran}
\bibliography{main}

\end{document}